\documentclass{article}

\usepackage[final]{nips_2016}

\usepackage[utf8]{inputenc} 
\usepackage[T1]{fontenc}    
\usepackage{hyperref}       
\usepackage{url}            
\usepackage{booktabs}       
\usepackage{amsfonts}       
\usepackage{nicefrac}       
\usepackage{microtype}      
\usepackage{multirow}
\usepackage{graphicx}
\usepackage{mathtools}
\usepackage{subcaption}

\title{Listen and Translate: A Proof of Concept for End-to-End Speech-to-Text Translation}

\author{
    Alexandre Bérard \thanks{also at LIG, Univ. Grenoble Alpes} \hspace{0.3cm} Olivier Pietquin \thanks{now with Google DeepMind - London, UK}\\
 Univ. Lille, CNRS, Centrale Lille, Inria, UMR 9189 CRIStAL - Lille, France\\
  \texttt{alexandre.berard@ed.univ-lille1.fr} \\ 
  \texttt{olivier.pietquin@univ-lille1.fr}
  \vspace{-0.5cm}
  \And
  Laurent Besacier \hspace{0.3cm} Christophe Servan \thanks{now with Systran, Paris, France} \\
  LIG, Univ. Grenoble Alpes, CNRS, Grenoble, France \\
  \texttt{firstname.lastname@imag.fr} \\
}
 
\begin{document}

\maketitle

\section{Introduction and background}

Current speech translation systems integrate (loosely or closely) two main modules: source language speech recognition (ASR) and source-to-target
text translation (MT).  In these approaches, source language text transcript (as a sequence or as a graph) appears as mandatory to produce a text hypothesis in the target language.
In the meantime, deep neural networks have yielded breakthroughs in different areas including machine translation and speech recognition. Current systems have an ``encoder-decoder'' architecture where a sequence of input symbols (or an input signal) is projected into a continuous low dimensional space and the output sequence is generated from this representation. This type of architecture has been proposed for machine translation \citep{sutskever2014sequence,bahdanau2014neural}
 and for automatic speech recognition \citep{chorowski2015attention,chan2015listen,Zweig2016}.
If we exclude some pioneering attempts to directly translate from a source phone sequence to a target word sequence \citep{BesacierZG06} or to discover a bilingual lexicon from parallel phone-word sequences \citep{godard:hal-01350119}, the only attempt to translate directly a source speech signal into target language text is that of \citet{duongattentional}. However, the authors focus on the alignment between source speech utterances and their text translation without proposing a complete end-to-end translation system. The same authors \citep{anastasopoulos2016unsupervised} also propose to jointly use IBM translation models (IBM Model 2), and dynamic time warping (DTW) to align source speech and target text, but again, only the alignment performance is measured in their work. 

Consequently, this paper proposes a first attempt to build an end-to-end speech-to-text translation system, which does not use source language text during learning or decoding.
Relaxing the need for source language transcription would drastically change the data collection methodology in speech translation, especially in under-resourced scenarios. For instance, in the former project DARPA TRANSTAC (speech translation from spoken Arabic dialects), a large effort was devoted to the collection of speech transcripts (and a prerequisite to obtain transcripts was often a detailed transcription guide for languages with little standardized spelling). Now, if end-to-end approaches for speech-to-text translation are successful, one might consider collecting data by asking bilingual speakers to directly utter speech in the source language from target language text utterances. Such an approach has the advantage to be applicable to any unwritten (source) language.
  
\section{Model}

\paragraph{Description}
We compare two end-to-end models: one for standard machine translation, and one for speech translation. Both models are attention-based encoder-decoder neural networks \citep{sutskever2014sequence,bahdanau2014neural}.

Given an input sequence $(x_1,\cdots,x_A)$, the encoder (a multi-layered bidirectional LSTM) produces a sequence of outputs $\mathrm{h}=(h_1,\cdots,h_A)$.
We initialize the decoder state with $s_0=tanh(W_{init}s^\prime_{A})$ where $s^\prime_{A}$ is the last state of the encoder. The next states of the decoder are computed as follows:
\begin{equation}
    s_t, o_t = LSTM(s_{t-1}, Ez_{t-1})
\end{equation}
$s_t$ is the new state of the LSTM, while $o_t$ is its output.
During training, $z_{t-1}$ is the previous symbol in the target (reference) sequence. During evaluation, $z_{t-1}$ is the predicted symbol at the previous time-step. $E$ is the target embedding matrix.

We compute an attention vector $d_t$ (over the encoder output $\mathrm{h}$), which is concatenated to the LSTM's output $o_t$:
\begin{equation}
    y_t = W_{proj}(o_t \oplus d_t)+b_{proj} \qquad d_t = attention(\mathrm{h}, s_t)
\end{equation}

This vector is then mapped to a vector of the same size as the target vocabulary, to compute a probability distribution over all target words.
At each time-step $t$, the greedy decoder picks the word $w_t$ with the highest probability according to this distribution.
\begin{equation}
    p(w_t|y_{t-1},s_{t-1}) = softmax(W_{out}y_t+b_{out})
\end{equation}

\vspace{-0.2cm}
\paragraph{Attention mechanism}
For text input, we use the same attention model as \citet{vinyals2015grammar}:
\vspace{-0.2cm}
\begin{equation}
    attention(\mathrm{h}, s_t) = \sum_{i=1}^{A}a_i^t h_i
\end{equation}
\vspace{-0.2cm}
\begin{equation}
    a_i^t = softmax(u_i^t) \qquad  u_i^t = v^Ttanh(W_1h_i+W_2s_t + b_2)
\end{equation}

\vspace{-0.2cm}
\paragraph{Convolutional attention}
For speech input, we use the same attention model as \citet{chorowski2015attention},
which uses a convolution filter $F$ to take into account the attention weights at the previous time-step.
This allows the attention model to ``remember'' and not translate the same part of the signal twice.
This is especially useful for speech as the input sequence can be quite long.
\begin{equation}
    u_i^t = v^Ttanh(W_1h_i+W_2s_t + b_2 + f_{t,i}\mu) \qquad f_t = F \ast a_{t-1}
\end{equation}

\vspace{-0.1cm}
$W_{init}\in\mathbb{R}^{2m\times 2m}$,
$E\in\mathbb{R}^{n\times \bar{V}}$,
$W_{proj}\in\mathbb{R}^{m\times 2m}$,
$W_{out}\in\mathbb{R}^{\bar{V}\times m}$,
$W_1\in\mathbb{R}^{m\times m}$,
$W_2\in\mathbb{R}^{m\times 2m}$,
$\mu\in\mathbb{R}^{m}$,
$F\in\mathbb{R}^{k}$,
where $n$ is the embedding size, $m$ the size of the LSTM layers (the state size is $2m$), $k$ the size of the convolution filter, and $\bar{V}$ the target vocabulary size.

\vspace{-0.2cm}
\paragraph{Implementation}
Our implementation is based on the \emph{seq2seq} model implemented by \emph{TensorFlow} \citep{tensorflow2015-whitepaper}. It reuses some of its components, while adding a number of features, like a bidirectional encoder  \citep{bahdanau2014neural}, a beam-search decoder, a convolutional attention model and a hierarchical encoder \citep{chorowski2015attention}.\footnote{Our source code is available online at \url{github.com/eske/seq2seq}}
Our implementation also allows pre-training, multiple encoders \citep{zoph2016multi} and multi-task training \citep{luong2015multi}.
Contrary to the reference implementation by \emph{TensorFlow}, which uses a bucket-based method with static (unrolled) RNNs, we use symbolic loops. This makes model creation much faster, and removes the need to manually configure bucket sizes.

\vspace{-0.2cm}
\paragraph{Model settings}
We describe two models: one for text translation, and one for speech translation. 
Both models use a decoder with 2 layers of 256 LSTM units, with word embeddings of size 256.
The \emph{speech} model does not have source embeddings, as the input is already a sequence of vectors (of 40 MFCCs + frame energy).
Both encoders are multi-layer bidirectional LSTMs, with 256 units. The \emph{text} model's encoder has 2 layers.
The \emph{speech} model has a hierarchical encoder with 3 layers, i.e., like \citet{chorowski2015attention}, the 2\textsuperscript{nd} and 3\textsuperscript{rd} layers read every other output from the previous layer, resulting in a sequence of 1/4th the size of the input. For the \emph{text} model, we use a standard attention model, while the \emph{speech} model uses convolutional attention (with a single filter of size 25). The latter also has two fully connected layers of size 256 between the input and the first layer of the encoder.

For training, we use the \emph{Adam} algorithm with an initial learning rate of $0.001$ \citep{kingma2014adam}, and a mini-batch size of $64$. We apply dropout (with a rate of $0.5$) during training on the connections between LSTM layers in the encoder and decoder \citep{Zaremba2014recurrent}.


\section{Experimental Settings}
\vspace{-0.2cm}

\begin{table}
\centering
\begin{tabular}{ | c | c | c | c | c | c | c | c | }
    \hline
    \multirow{2}{*}{Corpus} & \multirow{2}{*}{Lines} & \multicolumn{4}{c |}{French}  & \multicolumn{2}{c |}{English} \\
    \cline{3-8}
               &       & Words & Chars & Frames & Duration (h) & Words & Chars \\
    \hline
    BTEC train & 19972 & 201k  & 1020k & 5715k  & 15:51 & 189k  & 842k  \\
    BTEC dev   & 1512  & 12.2k & 61.9k & 357k   & 0:59  & 11.5k & 50.3k \\
    BTEC test1 & 469   & 3.9k  & 19.5k & 111k   & 0:18  & 3.6k  & 15.6k \\
    BTEC test2 & 464   & 3.8k  & 19.2k & 110k   & 0:18  & 3.6k  & 15.8k \\
    \hline
\end{tabular}
\vspace{0.2cm}
\caption{On the full corpus, there is on average 9.8 words/sent on the French side, and 9.3 on the English side.
    On the French audio side (with \emph{Agnes} as a speaker), there is on average 281 frames/sent (2.8 sec/sent).}
\label{table:BTEC}
\end{table}

As we do not have any good-sized speech translation corpus, we generated one with speech synthesis. 
We decided to use the French$\rightarrow$English BTEC corpus, which is a small parallel corpus (see Table \ref{table:BTEC}).
BTEC (Basic Travel Expression Corpus) is very clean and contains short sentences (10 words/sentence on average), with a small vocabulary (9218 unique tokens on the French side, 7186 on the English side). This corpus also has multiple references (7 references per source sentence for \emph{test1} and \emph{test2}), which we use for BLEU evaluation.  
We lowercase and tokenize all the data with the \emph{Moses} tokenizer.

Synthetic speech was generated using \emph{Voxygen}\footnote{\url{www.voxygen.fr}}, a commercial speech synthesis system, for 4 different female voices (Agnes, Marion, Helene, Fabienne) and 3 different male voices (Michel, Loic, Philippe). It is important to note that this is corpus-based concatenative speech synthesis 
\citep{Schwarz07a} and not parametric synthesis. So, for each speaker's voice, speech utterances are generated by concatenation of units mined from a large speech corpus (generally around 3000 sentences/speaker). This means that despite having very little intra-speaker variability in our speech data, there is a realistic level of inter-speaker variability. To challenge the robustness of our system to inter-speaker variability, we concatenate the training data for 6 speakers, and leave the last speaker (\textit{Agnes}) for evaluation only. 
 

Like \cite{chan2015listen}, as input for our \emph{speech} model, we segment the speech raw data into frames of 40 ms, with a step-size of 10 ms, and extract 40 MFCCs for each frame, along with the frame energy.

\begin{table}
\centering
\begin{tabular}{ | c | c | c | c | c || c | }
    \hline
    \multirow{2}{*}{Corpus} & \multicolumn{5}{c |}{BLEU score} \\
    \cline{2-6}
                    & Greedy & +BeamSearch & +LM     & +Ensemble & SMT \\
    \hline
    dev             & 42.5  & 43.6       & 45.1   & 51.6     & 54.3        \\
    \hline
    test1           & 39.8  & 41.7       & 43.0   & \textbf{47.1} & \textbf{49.4} \\
    \hline
    test2           & 41.0  & 41.0       & 42.8   & \textbf{47.9} & \textbf{45.8} \\
    \hline
    \hline
    test1 (7 refs)  & 50.7  & 53.4       & 54.7   & \textbf{60.3} & \textbf{60.0} \\
    \hline
    test2 (7 refs)  & 47.0  & 47.3       & 49.6   & \textbf{55.3} & \textbf{53.2} \\
    \hline
    \hline
    train.1000      & 78.4  & 80.5       & 82.9   & 89.6          & 76.6        \\
    \hline
\end{tabular}
\vspace{0.2cm}
    \caption{Results of the Text Translation task on BTEC, under different settings of the decoder. ``Greedy'' uses a greedy decoder.
    ``BeamSearch'' adds a beam-search decoder with a beam size of $8$. ``LM'' adds an external trigram language model estimated on the target side of \emph{BTEC train}. ``Ensemble'' uses a log-linear model with 5 NMT models trained independently (+ language model).
    The \emph{train} corpus used in evaluation is a subset of 1000 sentences from \emph{BTEC train}.
    For all the non-ensemble configurations, the best model out of 5 is used (according to its score on \emph{dev}). The SMT baseline is a phrase-based model (Moses) trained on \emph{BTEC train}, and tuned on \emph{BTEC dev}.
    }
    \label{table:text_results}
\end{table}

\begin{table}
\centering
\begin{tabular}{ | c | c | c | c | c | c || c | }
    \hline
    \multirow{2}{*}{Corpus} & \multirow{2}{*}{Speaker} & \multicolumn{5}{c |}{BLEU score} \\
    \cline{3-7}
                                    &                  & Greedy & +BeamSearch & +LM & +Ensemble & Baseline \\
    \hline
    \multirow{1}{*}{dev} & \multirow{3}{*}{\emph{Agnes}} & 30.1 & 32.3 & 33.5 & 40.2 & 43.7 \\
    \multirow{1}{*}{test1}          &                    & 29.2 & 31.9 & 32.7 & \textbf{38.6} & \textbf{42.1} \\
    \multirow{1}{*}{test2}          &                    & 29.0 & 30.6 & 31.0 & \textbf{37.1} & \textbf{39.6} \\
    \hline
    \hline
    \multirow{2}{*}{test1 (7 refs)} & \emph{Agnes}       & 37.6 & 40.1 & 41.8 & \textbf{48.8} & \textbf{52.7} \\
                                    & Michel             & 40.6 & 43.1 & 44.4 & \textbf{50.4} & \textbf{51.9} \\
    \hline
    \multirow{2}{*}{test2 (7 refs)} & \emph{Agnes}       & 33.6 & 35.6 & 36.7 & \textbf{43.1} & \textbf{46.7} \\
                                    & Michel             & 37.3 & 39.6 & 41.0 & \textbf{46.7} & \textbf{46.4} \\
    \hline
    \hline
    \multirow{1}{*}{train.1000}     & Michel             & 53.8 & 60.8 & 61.8 & 82.6 & 60.5 \\
    \hline
\end{tabular}
\vspace{0.2cm}
    \caption{Results of the speech translation experiments on BTEC. 
    The models are trained with 6 different speakers (including \emph{Michel}). \emph{Agnes} is not used for training.
    The \emph{Ensemble} configuration uses 5 models trained independently.
    The baseline system uses a pipeline Google Speech ASR + SMT system trained on BTEC (we trained a new system on French without punctuation, as Google Speech API does not output any punctuation symbol, except when actually pronounced).
    The WER scores obtained by the baseline ASR system range between $23\%$ and $26\%$.
    The last column from Table \ref{table:text_results} provides a baseline assuming a perfect ASR.
    } 
    \label{table:speech_results}
\end{table}

\vspace{-0.2cm}
\section{Experiments}
Table \ref{table:text_results} shows the results of our Machine Translation experiments. The main reason why we perform these experiments, is to assess whether it makes sense to train NMT models on such a small and specialized corpus.
Our best system (an ensemble of 5 models with a language model) gives similar results in terms of BLEU score to a baseline SMT system. This is rather surprising, considering the small amount of data used. Also, even though the models seem to overfit quite a lot (see \emph{train.1000} scores), this does not seem to be a problem during evaluation.

Table \ref{table:speech_results} shows the results of our Speech Translation experiments. Our best results are behind, but not very far from those of the baseline (which uses Google Speech API and a specialized SMT system). The results on a new speaker (\textit{Agnes}) are also encouraging, considering that we did not use any speaker adaptation technique.

\begin{figure}
\vspace{-0.7cm}
\centering
    \begin{subfigure}[b]{.5\textwidth}
    \centering
    \includegraphics[width=0.9\textwidth]{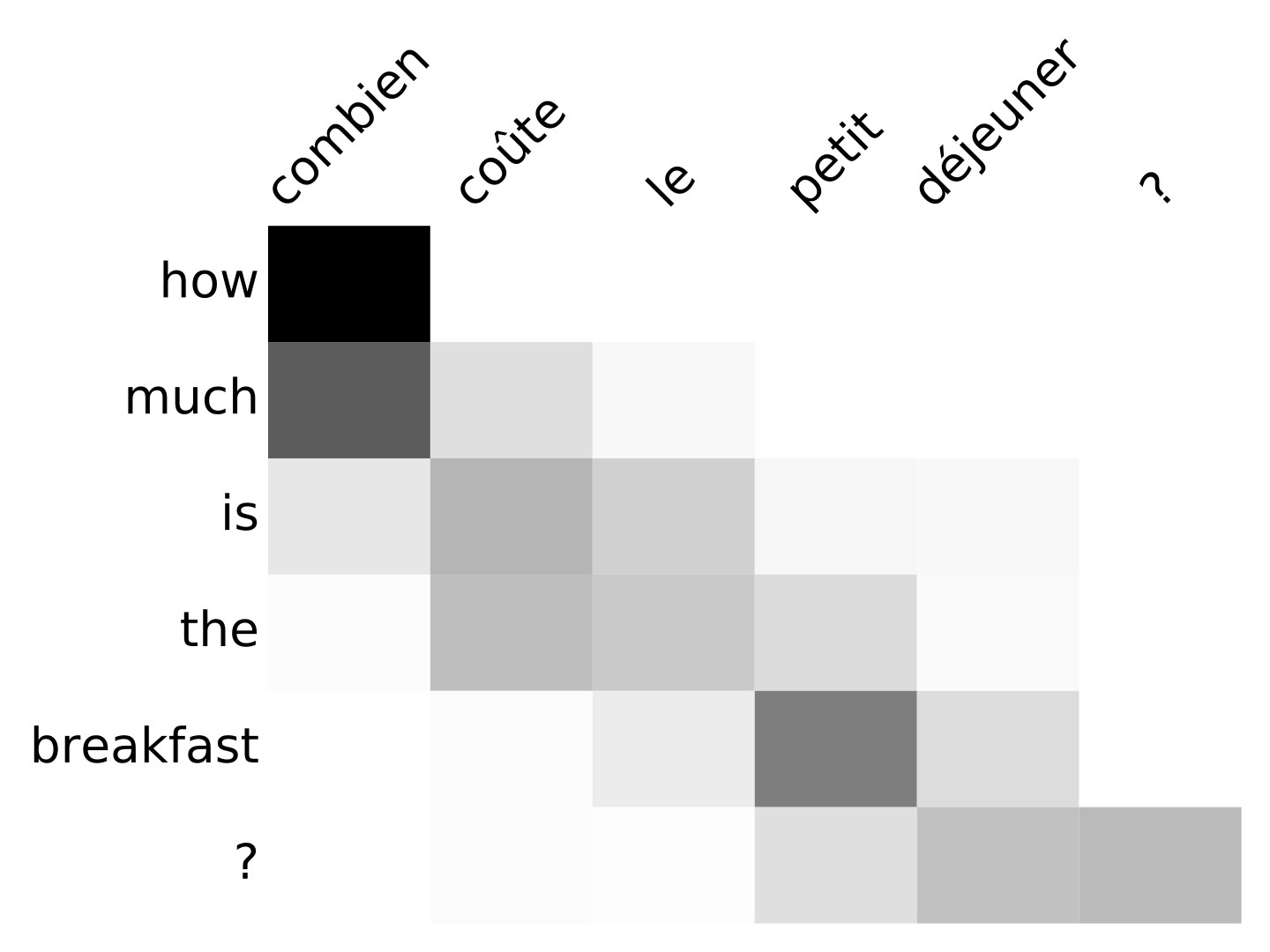}
    \caption{Machine translation alignment}
    \label{fig:text_alignment}
\end{subfigure}%
    \begin{subfigure}[b]{.5\textwidth}
    \centering
    \includegraphics[width=0.9\textwidth]{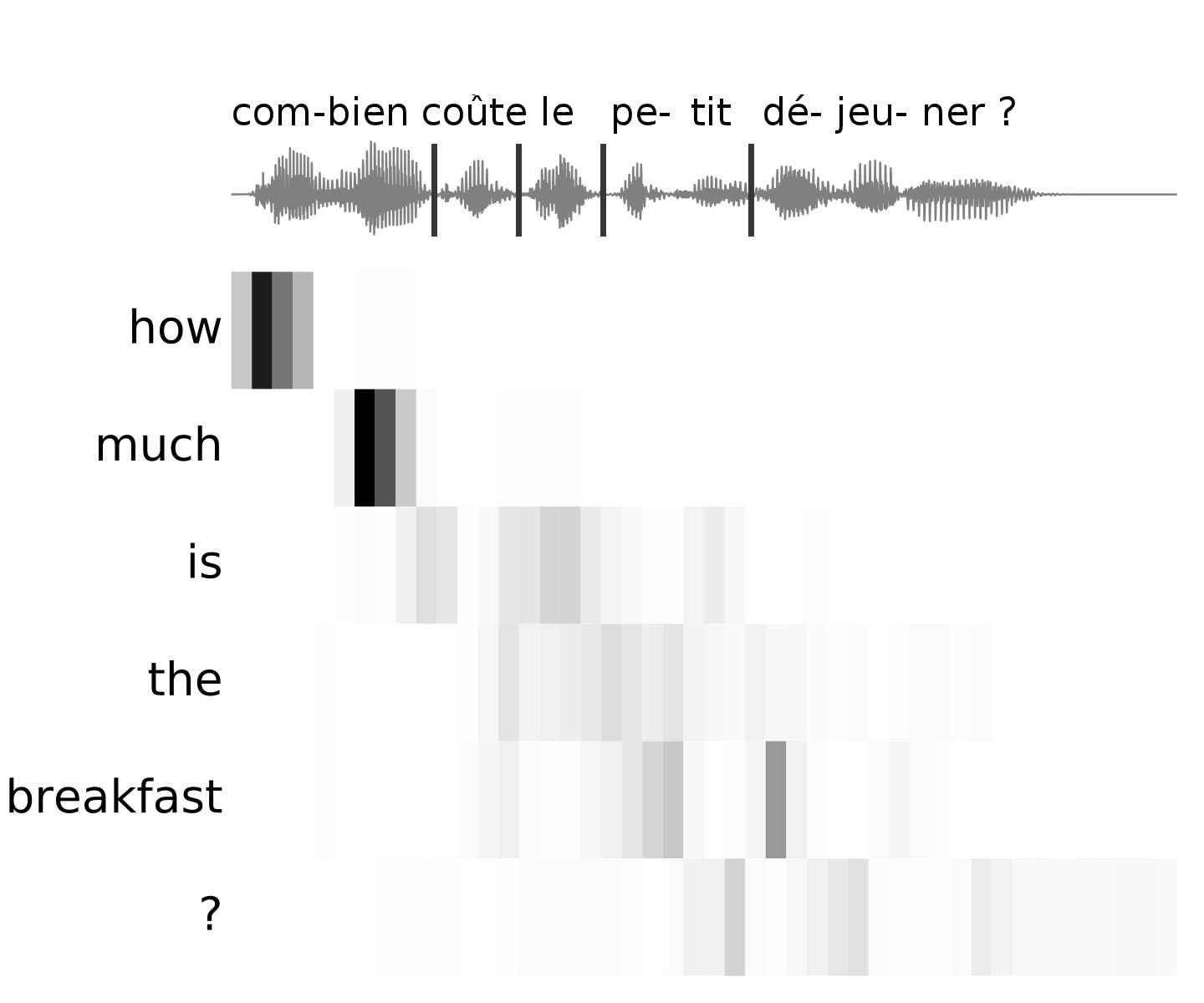}
    \caption{Speech translation alignment}
    \label{fig:speech_alignment}
\end{subfigure}
    \caption{Alignments performed by the attention model during training}
    \label{fig:alignment}
\vspace{-0.3cm}
\end{figure}

Figure \ref{fig:alignment} shows an example of alignment performed by the attention mechanism during training, for text and for speech.
We trained our models for $20k$ steps, which takes less than 2 hours for the \emph{text} models, and roughly 8 hours for the \emph{speech} models (on a single GTX 1070).

\vspace{-0.2cm}
\section{Conclusion}
We proposed a model for end-to-end speech translation, which gives promising results on a small French-English synthetic corpus.
These results, however, are to be mitigated due to the synthetic nature of the data: there is not much variability (except across speakers), which makes it easier for the
network to learn.
In any case, the experiments show that our architecture is able to generalize quite well to a new speaker.

Future work includes the evaluation of this model on non-synthetic data. Some datasets that could be used are: TED talks (there exists a parallel corpus and a speech corpus, but their alignment is not straightforward);
or books from the Gutenberg project\footnote{\url{www.gutenberg.org}} (public-domain books in many languages, along with some human-read audiobooks). We also wish to evaluate this model in a multi-source setting: does it help when translating text to have access to a human-read version of it, or other kinds of information such as part-of-speech or images?
 
 
\newpage
\bibliographystyle{apalike}
\bibliography{main}
\end{document}